\documentclass{article}

\usepackage[final]{iai_neurips_2024}

\usepackage[utf8]{inputenc} 
\usepackage[T1]{fontenc}    
\usepackage{hyperref}       
\usepackage{url}            
\usepackage{booktabs}       
\usepackage{amsfonts}       
\usepackage{nicefrac}       
\usepackage{microtype}      
\usepackage{xcolor}         

\usepackage{amsmath}
\usepackage{makecell}
\usepackage{multirow}
\usepackage{caption}
\usepackage{array, longtable}
\usepackage{csquotes}
\usepackage{float}
\usepackage[flushleft]{threeparttablex}

\usepackage{dblfloatfix}
\usepackage{graphicx}
\usepackage{svg}
\graphicspath{ {./figures/} }
\svgsetup{inkscapepath=svgsubdir}

\usepackage{algorithm}
\usepackage{algpseudocode}

\usepackage{pgffor}
\usepackage{ifthen}

\title{Clustering and Alignment: Understanding the Training Dynamics in Modular Addition}

\author{
  Tiberiu Mușat \\
  ETH Zürich \\
  \texttt{tmusat@ethz.ch} \\
}

\newboolean{showfigures}
\setboolean{showfigures}{true} 

\begin{document}

\maketitle

\begin{abstract}
    Recent studies have revealed that neural networks learn interpretable
    algorithms for many simple problems. However, little
    is known about how these algorithms emerge during training. In this article,
    I study the training dynamics of a small neural network with 2-dimensional embeddings
    on the problem of modular addition. I observe that embedding vectors tend to organize into two types of
    structures: grids and circles. I study these structures and explain
    their emergence as a result of two simple tendencies exhibited by pairs of
    embeddings: clustering and alignment. I propose explicit formulae for these
    tendencies as interaction forces between different pairs of embeddings.
    To show that my formulae can fully account for the emergence of these structures,
    I construct an equivalent particle simulation where I show that identical structures emerge.
    I discuss the role of weight decay in my setup and reveal a new mechanism
    that links regularization and training dynamics. To support my findings, I also release an interactive demo
    available at \url{https://modular-addition.vercel.app/}.
\end{abstract}

\section{Introduction}

Mechanistic interpretability aims at reverse-engineering the inner workings of
trained neural networks and explaining their behavior in terms of interpretable
algorithms. Recent work in this field was very successful in uncovering the
algorithms learned by neural networks on simple problems.
\citet{zhong2023clock} showed that transformers learn to solve modular addition
by forming circular structures in the embedding space and applying one of two
simple algorithms: a ``Clock'' algorithm (resembling the way humans read the
clock) and a ``Pizza'' algorithm (unfamiliar, but interpretable; also
encountered in this article). \citet{charton2024gcd} showed that transformers
learn to compute the greatest common divisor by identifying the prime factors
of the numeric base from the last digits of each number.
\citet{quirke2024understanding} uncovered that transformers break down the
multi-digit addition task into parallel, digit-specific streams, using
different algorithms for various digit positions.

However, it remains an open problem to provide a similarly high degree of
interpretability for the training dynamics that lead to the emergence of these
algorithms. Currently, the most successful approach to understanding the
learning process is by uncovering hidden progress measures that increase
abruptly during training \citep{barak2022hidden}. For example, by constructing
two progress measures for the problem of modular addition,
\citet{nanda2023progress} find that training of neural networks on modular
addition can be split into three phases: memorization, circuit formation, and
cleanup. While insightful, such methods provide only a high-level description
of the training process, without offering a clear explanation of the underlying
optimization dynamics. A better understanding of neural networks could lead to
AI systems that are more interpretable, more efficient, and more reliable
\citep{doshivelez2017rigorous,olah2020zoom}.

\section{My contribution}

In this article, I study the training dynamics of a small neural network on the
problem of modular addition. My architecture consists in a simplified
single-layer transformer with constant attention and 2-dimensional embeddings.
My contributions are the following.

\paragraph*{Grids and circles}

\begin{figure}[h]
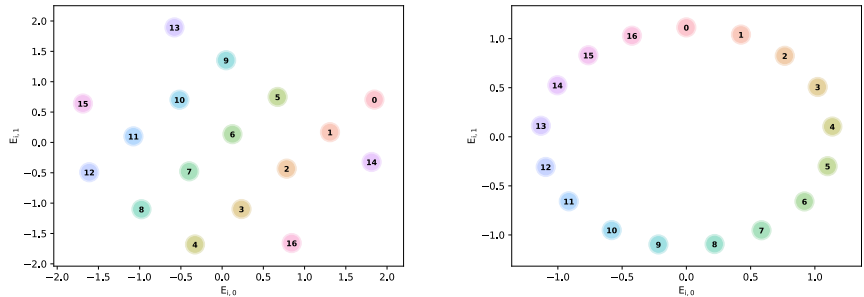

    \centering
    \includesvg[width=6cm]{svg/embeds_wd1.0/99.svg}
    \includesvg[width=6cm]{svg/embeds_wd1.0/80.svg}
    \caption{Embedding vectors self-organize into grids (left)
        and circles (right).}
    \label{fig:structures}
\end{figure}

I show that embedding vectors tend to organize themselves into circular and
grid-like structures, as depicted in Figure \ref*{fig:structures}. The
emergence of circular and grid-like structures is consistent with the findings
of \citet{zhong2023clock}, \citet{gromov2023grokking} and
\citet{liu2022grokking}. I show that grids and circles play a key role in the
generalization performance of the model. I also show that weight decay plays a
crucial role in the emergence of circular structures and in reducing grid
imperfections.

\paragraph*{Clustering and alignment}

I show how grids and circles facilitate accurate classification for the
subsequent layers by grouping ``pair sums'' (outputs of the constant
attention). I propose an explanation for the emergence of the grids and circles
as a result of two simple phenomena: clustering and alignment. I propose
explicit formulae for both phenomena as interaction forces between two
different pairs of embeddings.

\paragraph*{Particle simulation}

To prove that my proposed formulae can fully account for the emergence of grids
and circles, I construct a particle simulation where particles correspond to
embedding vectors and forces correspond to gradients. Forces are computed using
only my proposed formulae for clustering and alignment. In this simulation, I
show that the particles self-organize into the same structures as the
embeddings in the trained transformer: circles, grids, and imperfect grids.

I also contribute to the understanding of weight decay by discussing its role
in my particular setup in Appendix \ref{appendix:weight_decay}. I also release
an interactive demo to allow the readers to explore the training dynamics and
particle simulations for themselves:
\url{https://modular-addition.vercel.app/}.

\section{Architecture}

In this article, I study a simplified single-layer transformer with constant
attention. A similar setup has been previously used by \citet{liu2022grokking},
\citet{zhong2023clock} and \citet{hassid-etal-2022-attention}.

The input to the model consists of two tokens $a$ and $b$ representing the two
numbers to be added, where $a, b \in \{0, 1, \ldots, N - 1\}$. The tokens are
embedded into vectors $x_a$ and $x_b$ using an embedding matrix $E \in
    \mathbb{R}^{N \times D}$, where $D$ is the embedding dimension. For the scope
of this article, I focus on the case when $D = 2$ for simplicity and ease of
visualization. Then, a constant attention mechanism is applied to the
embeddings. This is equivalent to computing the sum of the embedding vectors.

Then I apply a linear layer of size $H$ with weight matrix $W_h \in
    \mathbb{R}^{H \times D}$, bias vector $b_h \in \mathbb{R}^H$, and ReLU
activation. Finally, I apply a linear layer of size $N$ with a weight matrix
$W_o \in \mathbb{R}^{N \times H}$ and bias vector $b_o \in \mathbb{R}^N$. The
output of the model is the logits of the predicted sum. I don't use any skip
connections or normalization layers.

We can formalize the model as follows:

\begin{itemize}
    \item Inputs: $a, b \in \{0, 1, \ldots, N - 1\}$
    \item Embeddings: $x_a = E_a$, $x_b = E_b$
    \item Constant attention: $x = x_a + x_b$
    \item Linear layer: $h = \mathrm{ReLU}(W_h x + b_h)$
    \item Output layer: $o = W_o h + b_o$
\end{itemize}

I train the model in full batch mode using the Adam optimizer with a learning
rate of $0.01$ and decoupled weight decay \citep{loshchilov2019adamw} between
$0$ and $1$. I use the cross-entropy loss. Unless specified, I use $N = 17$ and
$H = 32$. The training set consists of 80\% of all $N(N+1)/2$ distinct pairs of
numbers, chosen randomly. The validation set consists of the remaining 20\%.

\begin{figure}[h]
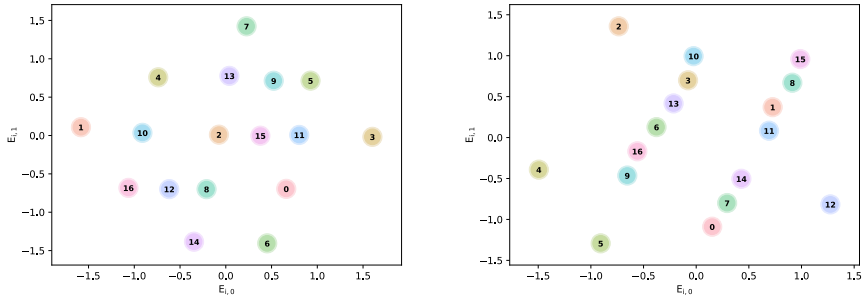

    \centering
    \includesvg[width=6cm]{svg/embeds_wd1.0/32.svg}
    \includesvg[width=6cm]{svg/embeds_wd1.0/77.svg}
    \caption{Sometimes embedding vectors self-organize into imperfect grids.}
    \label{fig:imperfect_grids}
\end{figure}

\section{Grids and circles: the key to successful generalization}
\label{sec:structures}

In trained models, the embedding vectors tend to be positioned in arithmetic
progressions, forming either grids or circles. I visualize two examples of
these structures in Figure \ref{fig:structures}. The validation accuracy is
highest when the embeddings form these structures most clearly. Sometimes,
however, the embeddings self-organize into imperfect grids, as shown in Figure
\ref{fig:imperfect_grids}. In such cases, the validation accuracy is lower, but
still higher than when the embeddings are not aligned at all. We present more
examples of these structures in Appendix \ref{appendix:embeddings}.

To quantify this effect, I devise two simple algorithms: one to detect circles,
and another to quantify the number of grid imperfections in non-circular
structures. Both algorithms are explained in detail in the Appendix
\ref{appendix:algorithms}.

\begin{table}[h]
    \centering
    \begin{threeparttable}
        \caption{Validation accuracy and structures formed by embedding vectors.}
        \begin{tabular}{ccccccc}
            \toprule
                         & \multicolumn{2}{c}{Circles} & \multicolumn{4}{c}{Non-Circles}                                                          \\

            \cmidrule(r){2-3}
            \cmidrule(r){4-7}
            Weight Decay & Num                         & Acc                             & Num & Acc  & Grid Imperfections & Correlation          \\
            \midrule
            0.0          & 0                           & -                               & 100 & 18.2 & $102.2 \pm 6.1$    & -0.92 [-0.95, -0.88] \\
            0.3          & 1                           & 77.4                            & 99  & 34.8 & $72.7 \pm 6.6$     & -0.90 [-0.93, -0.85] \\
            0.6          & 12                          & 65.9                            & 88  & 51.6 & $39.6 \pm 4.0$     & -0.84 [-0.89, -0.77] \\
            1.0          & 18                          & 60.4                            & 82  & 46.7 & $31.6 \pm 4.0$     & -0.64 [-0.75, -0.49] \\
            \bottomrule
        \end{tabular}

        \begin{tablenotes}
            \small
            \item\hskip -\fontdimen2{Explanation}: I train 100 random initializations for 2000 epochs for various values of weight decay. \\ For each value of weight decay, I report the following: the number of circles and their average validation accuracy, the number of non-circles and their average validation accuracy, the average number of grid imperfections for non-circles, and the correlation between the number of grid imperfections and the validation accuracy of non-circles. For circle detection and grid imperfections, see Appendix \ref{appendix:algorithms}.
        \end{tablenotes}
        \label{table:structures}
    \end{threeparttable}
\end{table}

I run multiple experiments for various values of weight decay and measure the
average validation accuracy of circles and non-circles, as well as the
correlation between the validation accuracy and the number of grid
imperfections of non-circles. I find that circles consistently have higher
validation accuracy than non-circles. For non-circles, I find a strong negative
correlation between the number of grid imperfections and the validation
accuracy.

\section{Co-evolution of embeddings and linear layers}
\label{sec:co-evolution}

In this section, I will model the training process as the co-evolution of two
systems: the embeddings and the subsequent layers. The constant attention
mechanism sums the embedding vectors to obtain $x = E_a + E_b$. The subsequent
layers then act as a classifier, trying to classify this value as $c = a + b\
    (\mathrm{mod}\ N)$ out of $N$ possible values.

\begin{figure}[h]
    \centering
    \captionsetup{justification=centering}
    \includesvg[height=5cm]{svg/color_map/03.svg}
    \includegraphics[height=5cm]{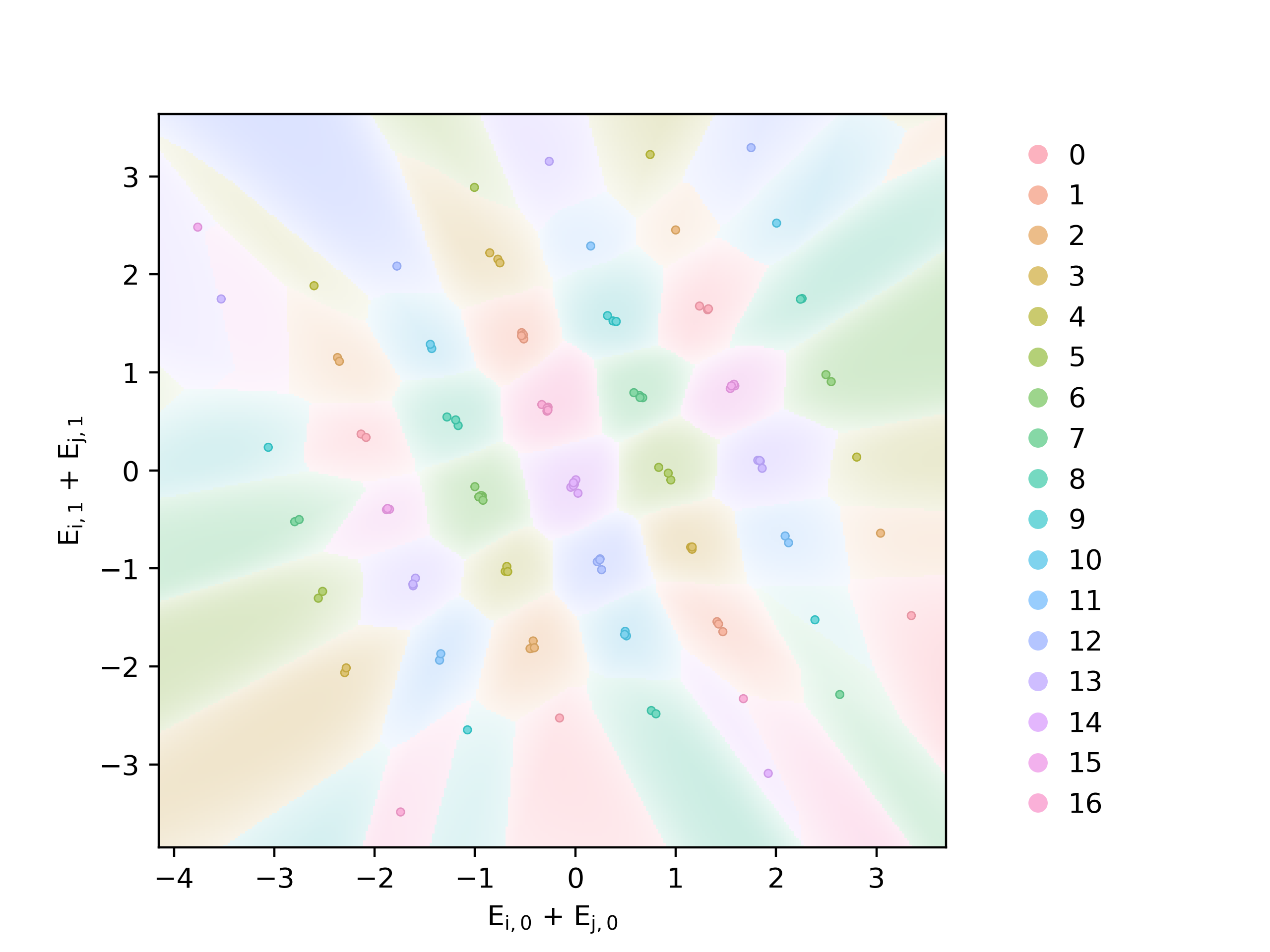}
    \caption{Embedding vectors (left); classifier formed by the combined linear and output \\ layers (background, right) and sums of embedding pairs in the training set (markers, right).}
    \label{fig:classifier_grid}
\end{figure}

\subsection{Clustering}

As depicted in Figure \ref{fig:classifier_grid}, the pair sums tend to cluster
based on their modular sum. I present more examples in Appendix
\ref{appendix:classifier}. Let's try to understand this phenomenon by
considering the interaction between two pair sums $x_{ij} = E_i + E_j$ and
$x_{kl} = E_k + E_l$, where $i, j, k, l \in \{0, 1, \ldots, N - 1\}$, and $(i,
    j)$ and $(k, l)$ are two distinct pairs of numbers in the training set.

Let's first consider the case when $i + j \not \equiv k + l\ (\mathrm{mod}\
    N)$. For simplicity, let's assume that there are no other pair sums in between
$x_{ij}$ and $x_{kl}$. In this case, to classify $x_{ij}$ and $x_{kl}$
correctly, the classifier must place a decision boundary between them. This
decision boundary will create a gradient that will push $x_{ij}$ and $x_{kl}$
away from each other. Moreover, the closer $x_{ij}$ and $x_{kl}$ are, the
narrower the decision boundary will have to be, resulting in a stronger
gradient and a stronger push. In other words, the push will be inversely
proportional to the distance between the pair sums. I propose the following
formula for the gradient at $x_{ij}$ induced by the pair sum $x_{kl}$:

\begin{equation} \label{eq:clustering_repulsion}
    g_{ij \leftarrow kl} = g_{r} \cdot \frac{x_{ij} - x_{kl}}{\| x_{ij} - x_{kl} \|} \cdot \frac{1}{\| x_{ij} - x_{kl} \|}
\end{equation}

where $g_{r}$ is a repulsion constant. The gradient at $x_{kl}$ is defined
analogously: $g_{kl \leftarrow ij} = -g_{ij \leftarrow kl}$.

Now, let's consider the case when $i + j \equiv k + l\ (\mathrm{mod}\ N)$.
Again, for simplicity, let's assume that there are no other pair sums in
between $x_{ij}$ and $x_{kl}$. In this case, $x_{ij}$ and $x_{kl}$ are likely
to be inside the same classification region. Both pair sums will be attracted
towards the point of maximum classification accuracy, which on average will be
the center of the classification region. I will model this case with a constant
gradient towards the other pair sum:

\begin{equation} \label{eq:clustering_attraction}
    g_{ij \leftarrow kl} = - g_{a} \cdot \frac{x_{ij} - x_{kl}}{\| x_{ij} - x_{kl} \|}
\end{equation}

where $g_{a}$ is an attraction constant. The gradient at $x_{kl}$ is defined
analogously: $g_{kl \leftarrow ij} = -g_{ij \leftarrow kl}$. Similar
interactions have been previously studied by \citet{baek2024geneft},
\citet{liu2022grokking}, and \citet{rossem2024representations}.

\subsection{Alignment}

\begin{figure}[h]
    \centering
    \captionsetup{justification=centering}
    \includesvg[height=5cm]{svg/color_map/06.svg}
    \includegraphics[height=5cm]{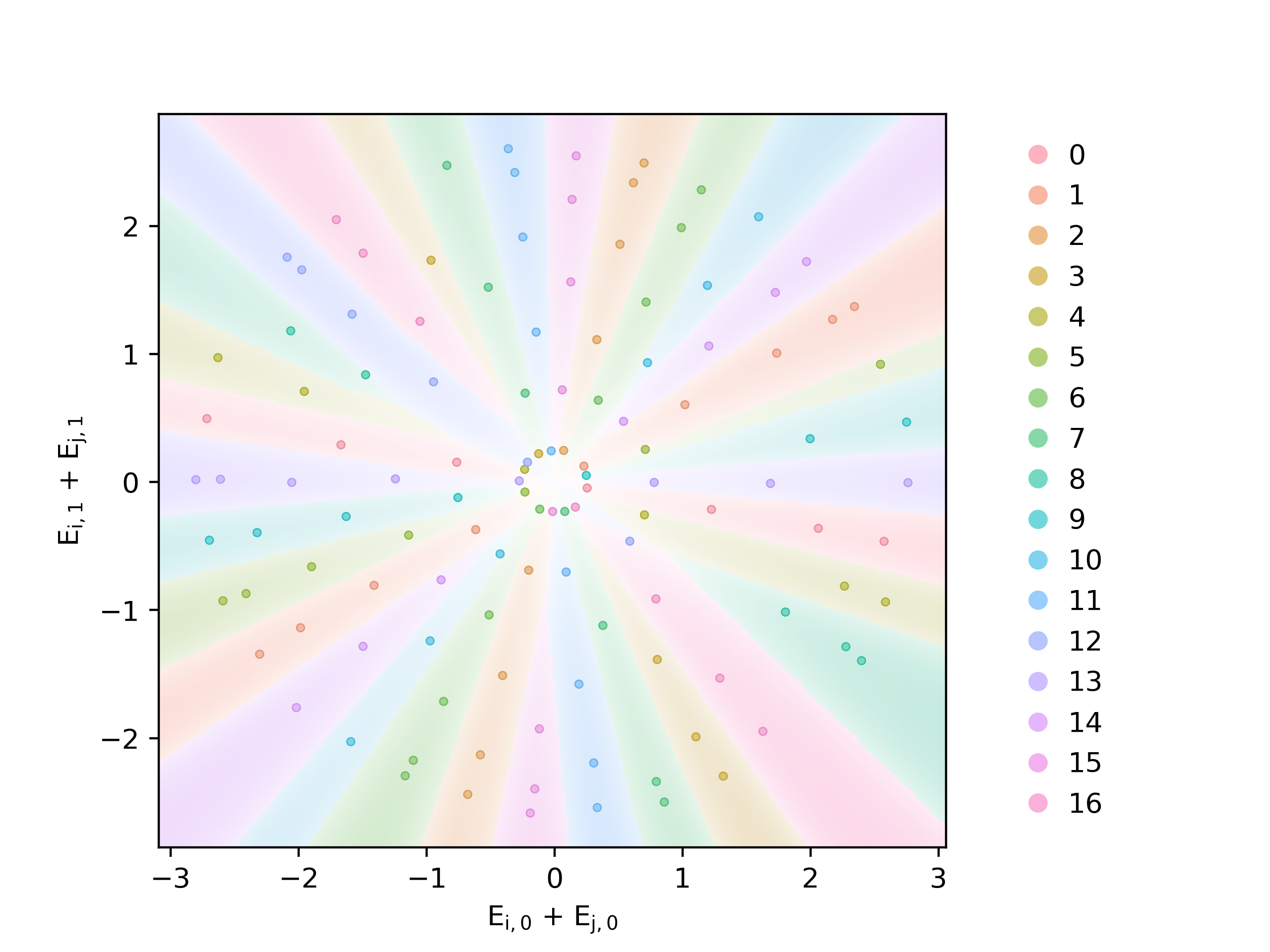}
    \caption{Embedding vectors (left); classifier formed by the combined linear and output \\ layers (background, right) and sums of embedding pairs in the training set (markers, right).}
    \label{fig:classifier_circle}
\end{figure}

A different picture emerges when we look at a model where the embeddings form a
circle. In Figure \ref{fig:classifier_circle}, I visualize the embedding
vectors, pair sums, and the classification function of a model with circular
embeddings at the end of training. Pair sums are no longer clustered. Rather,
they are aligned along lines that pass through the origin. In this arrangement,
the classification regions are skewed and start from the origin, resembling a
pizza. This model corresponds exactly to the ``pizza'' algorithm described by
\citet{zhong2023clock}.

In Table \ref{table:structures}, we saw that weight decay plays a crucial role
in the emergence of circular structures. To understand the mechanism behind the
alignment, we need to consider that the classification function is just a sum
of ReLU activations of the linear layer.

Weight decay encourages the linear layer to have small weights and biases,
which results in linear functions with small slopes and close to the origin. We
explore the precise impact of weight decay on the magnitude of the weights and
biases in Appendix \ref{appendix:magnitudes}. The only aspect that remains
completely unconstrained in the direction of the weight vectors. A weight
vector aligned with a specific pair sum is maximally useful for the
classification layer because it outputs the maximum possible value for that
pair sum and the minimum possible value for the other pair sums. Thus, under
these conditions, in order to minimize the training loss, the weight vectors of
the linear layer will tend to align with the pair sums.

We can model this behavior as follows. For a single pair sum $x_{kl} = E_k +
    E_l$, let's consider a linear function with zero bias ($b = 0$) and a weight
vector of constant norm that is perfectly aligned with the pair sum ($w = f_a
    \cdot \frac{x_{kl}}{\| x_{kl} \|}$, where $f_a$ is a constant). After applying
the ReLU activation, the output becomes zero for all points $x$ that oppose the
direction of $x_{kl}$ (i.e., $x \cdot x_{kl} < 0$). For the other points, the
function will have a constant slope of $f_a$ in the direction of $x_{kl}$.

Inspired by this simplified model, I propose the following formula for the
gradient at $x_{ij}$ produced by the pair sum $x_{kl}$:

\begin{equation} \label{eq:alignment}
    g_{ij \leftarrow kl} = \begin{cases}
        0,                                        & \text{if $x_{ij} \cdot x_{kl} \leq 0$}                                              \\
        f_{a} \cdot \frac{x_{kl}}{\| x_{kl} \|},  & \text{if $x_{ij} \cdot x_{kl} > 0$ and } i + j \equiv k + l\ (\mathrm{mod}\ N)      \\
        -f_{a} \cdot \frac{x_{kl}}{\| x_{kl} \|}, & \text{if $x_{ij} \cdot x_{kl} > 0$ and } i + j \not \equiv k + l\ (\mathrm{mod}\ N) \\
    \end{cases}
\end{equation}

\section{Particle simulation}

In this section, I show that my proposed model can fully account for the
emergence of the grids and circles by constructing a particle simulation and
showing that particles converge to the same structures as the embedding vectors
in the trained transformer.

\begin{figure}[h]
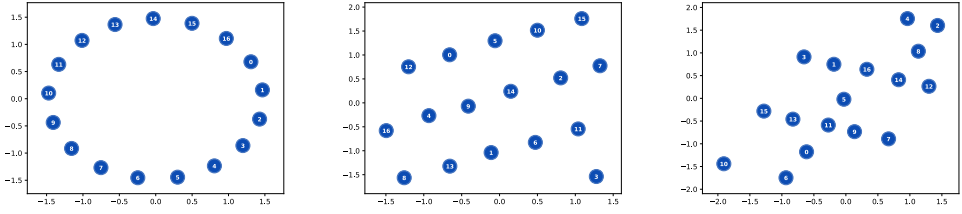

    \centering
    \captionsetup{justification=centering}
    \includesvg[height=3.3cm]{svg/particles_2.0/6.svg}
    \includesvg[height=3.3cm]{svg/particles_2.0/5.svg}
    \includesvg[height=3.3cm]{svg/particles_2.0/65.svg}
    \caption{Particles form circles (left), grids (center), and imperfect grids (right).}
    \label{fig:particle_sim}
\end{figure}

I model the training dynamics of the embedding vectors using $N$ particles in a
$D$-dimensinal space ($N = 17, D = 2$). I denote the position of particle $i$
by $x_i \in \mathbb{R}^D$, $i = 1, \ldots, N$. The position $x_i$ of particle
$i$ corresponds to the embedding vector $E_i$ of the number $i$ during
training.

The gradient equations from the previous section become ``forces'' acting on
the $N$ particles. Let $(i, j)$ and $(k, l)$ be two distinct pairs of numbers.
I denote the sums of their particle positions as $x_{ij} = x_i + x_j$ and
$x_{kl} = x_k + x_l$, respectively. The pair $(k, l)$ will induce an equal
force on particles $i$ and $j$ (obtained by combining equations
\ref{eq:clustering_repulsion}, \ref{eq:clustering_attraction}, and
\ref{eq:alignment}):

\[
    F_i = F_j = \begin{cases}
        g_a \cdot \frac{x_{kl} - x_{ij}}{\| x_{kl} - x_{ij} \|}                                            & \text{if $x_{ij} \cdot x_{kl} \leq 0$ and $i + j \equiv k + l\ (\mathrm{mod}\ N)$}      \\
        g_r \cdot \frac{x_{ij} - x_{kl}}{\| x_{ij} - x_{kl} \|^2}                                          & \text{if $x_{ij} \cdot x_{kl} \leq 0$ and $i + j \not \equiv k + l\ (\mathrm{mod}\ N)$} \\
        g_a \cdot \frac{x_{kl} - x_{ij}}{\| x_{kl} - x_{ij} \|} + f_{a} \cdot \frac{x_{kl}}{\| x_{kl} \|}  & \text{if $x_{ij} \cdot x_{kl} > 0$ and $i + j \equiv k + l\ (\mathrm{mod}\ N)$}         \\
        g_r \cdot \frac{x_{ij} - x_{kl}}{\| x_{ij} - x_{kl} \|^2} -f_{a} \cdot \frac{x_{kl}}{\| x_{kl} \|} & \text{if $x_{ij} \cdot x_{kl} > 0$ and $i + j \not \equiv k + l\ (\mathrm{mod}\ N)$}    \\
    \end{cases}
\]

The force acting on particles $k$ and $l$ is defined analogously. I use $g_r =
    g_a = f_a = 1$. Forces are weighted so that interactions between pair sums of
different modular sum have the same total contribution as those between pair
sums of the same modular sum. Particles are initialized randomly according to a
normal distribution. At every step of the simulation, the total force acting on
each particle is calculated. The, each particle is moved with a small step in
the direction of the total force. I also scale the particle positions to
maintain a constant variance and zero mean. I repeat this process for 100
steps. No momentum is used.

I observe that the particles self-organize into the same structures as the
embeddings in the trained transformer: circles, grids, and imperfect grids. We
visualize a few examples in Figure \ref{fig:particle_sim}. For more examples,
see Appendix \ref{appendix:particles}. In Table \ref{table:particle_sim}, we
present the results of running 100 simulations for various values of $f_a$. We
find that the relative frequency of circles and the average number of grid
imperfections are very similar to the results obtained from the transformer
experiments. I also find that increasing $f_a$ leads to more circles, which is
consistent with my hypothesis that circles emerge as a result of the alignment
force.

\begin{table}[h]
    \caption{Results of particle simulations for various values of $f_a$}
    \centering
    \begin{tabular}{ccccccc}
        \toprule
        $f_a$ & Number of Circles & Number of Grids & Average Grid Imperfections \\
        \midrule
        0.5   & 0                 & 100             & $35.5 \pm 3.7$             \\
        1.0   & 17                & 83              & $36.3 \pm 4.8$             \\
        2.0   & 56                & 44              & $42.9 \pm 10.1$            \\
        \bottomrule
    \end{tabular}
    \label{table:particle_sim}
\end{table}

\section{Conclusion}
\label{sec:conclusion}

I have explained the training dynamics of a simplified transformer on the
problem of modular addition in terms of two simple phenomena: clustering and
alignment. I have provided strong empirical evidence for my model using a
particle simulation. I have also provided a qualitative argument for the two
phenomena, but further work is needed to fully understand their origin.

\bibliography{iai_neurips_2024}


\newpage
\appendix

\section{Algorithms}
\label{appendix:algorithms}

\subsection{Circle detection}

To determine whether the embedding vectors form a circle, I simply consider the
ratio of the maximum and minimum distances of the embeddings to the origin. If
the ratio is below a certain threshold, I consider the embeddings to form a
circle. I use a threshold value of 1.2. For my setup, I observe that this
simple algorithm is sufficient to detect circles with great accuracy.

\begin{algorithm}
    \caption{Check if Embeddings Form a Circle}
    \begin{algorithmic}[1]
        \Function{is\_circle}{E: embedding matrix}
        \State $min\_norm \gets \min(\sqrt{x^2 + y^2} \mid (x, y) = E_i, i \in \{0, \ldots, N-1\})$
        \State $max\_norm \gets \max(\sqrt{x^2 + y^2} \mid (x, y) = E_i, i \in \{0, \ldots, N-1\})$
        \State \Return $max\_norm / min\_norm < 1.2$
        \EndFunction
    \end{algorithmic}
\end{algorithm}

\subsection{Grid imperfections}

I propose a simple measure to quantify the number of grid imperfections
inspired by the following fact: in a perfect grid, the vector sums of all pairs
of embeddings also form a perfect grid where pair sums are grouped together
based on their modular sum.

My measure works as follows. For each pair of embeddings $(E_i, E_j)$, I find
the pair of embeddings $(E_k, E_l)$ with the closest vector sum. If the modular
sum $(i + j)\ \mathrm{mod}\ N$ is different from the modular sum $(k + l) \
    \mathrm{mod}\ N$, I consider this pair to be an ``imperfection''.

\begin{algorithm}[h]
    \caption{Count Grid Imperfections in Embedding}
    \begin{algorithmic}[1]
        \Function{imperfections}{E: embedding matrix}
        \State $N \gets \text{len}(embed)$
        \State $pairs \gets \{(i, j) \mid i \in \{0, \ldots, N-1\}, j \in \{i, \ldots, N-1\}\}$
        \State $imperfections \gets 0$
        \ForAll{$(i, j) \in pairs$}
        \State $min\_dist \gets \infty$
        \State $match \gets \text{False}$
        \ForAll{$(k, l) \in pairs$}
        \If{$(i, j) = (k, l)$}
        \State \textbf{continue}
        \EndIf
        \State $dist \gets \| E_i + E_j - E_k - E_l \|$
        \If{$dist < min\_dist$}
        \State $min\_dist \gets dist$
        \State $match \gets i + j \equiv k + l \mod N$
        \EndIf
        \EndFor
        \If{\textbf{not} $match$}
        \State $imperfections \gets imperfections + 1$
        \EndIf
        \EndFor
        \State \Return $imperfections$
        \EndFunction
    \end{algorithmic}
\end{algorithm}

\newpage
\section{Weight decay and the magnitude of weights and biases}
\label{appendix:magnitudes}

\begin{figure}[h]
    \centering
    \captionsetup{width=0.7\textwidth, justification=centering}
    \includesvg[width=12cm]{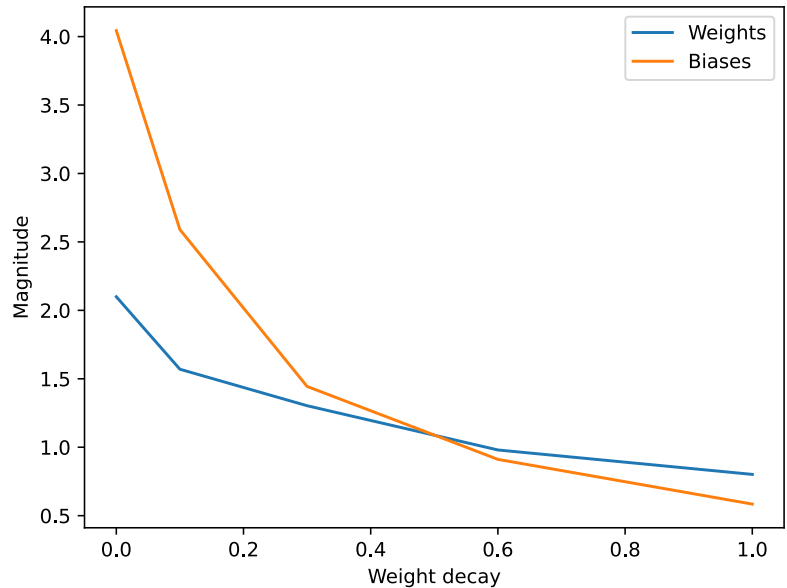}
    \caption{Average magnitude of the weights and biases of the linear layer at the end of training for various values of weight decay.}
    \label{fig:magnitudes}
\end{figure}

Below I visualize the magnitude of the weights and biases of the linear layer
at the end of training for various values of weight decay. I run $10$ random
initializations for $2000$ epochs with a learning rate of $0.01$. I plot the
average absolute value of the weights and biases of the linear layers among the
$10$ runs in Figure \ref{fig:magnitudes}. We can observe that weight decay
strongly limits the magnitude of the weights and biases, with biases being
limited more than weights.

\newpage
\section{The role of weight decay}
\label{appendix:weight_decay}

Weight decay is a broadly used regularization technique for training
state-of-the-art deep networks \citep{loshchilov2019adamw, krogh1991simple,
    andriushchenko2023weightdecay}. Weight decay penalizes large weights and biases
by applying an exponential decay to each parameter ($\theta$) at every step of
the optimization. The decay rate is proportional to the learning rate
($\lambda$) and the weight decay coefficient ($\gamma$), as shown in the
following formula:

\begin{equation}
    \theta_{t+1} \leftarrow \theta_{t} - \lambda \gamma \theta_{t}
\end{equation}

Traditionally, weight decay was understood as a form of $L_2$ regularization,
improving generalization by constraining the network capacity
\citep{Goodfellow-et-al-2016}. More recently, a new perspective on weight decay
has emerged, suggesting that it plays a much more important role during
training by changing the training dynamics in a desirable way
\citep{zhang2018three}. In this section, I combine the two perspectives by
discussing the impact of weight decay on the training dynamics in my setup.

In Section \ref{sec:structures}, we observed that weight decay plays an
important role in the successful generalization of the trained model. I propose
the following explanation: weight decay changes the training dynamics by
strengthening the clustering force and by introducing the alignment force. In
turn, these forces facilitate the emergence of grids and circles, which are
crucial for the generalization performance of the model. Below I discuss
exactly how weight decay impacts clustering and alignment.

In the case of clustering, my proposed clustering force (Equations
\ref{eq:clustering_repulsion} and \ref{eq:clustering_attraction}) is highly
dependent on a simplified model of the classification function where the
decision boundaries are very wide. Weight decay enables this by limiting the
magnitude of the weights of the linear layer and output layer, as I show in
Appendix \ref{appendix:magnitudes}. Without weight decay, classification
boundaries could get arbitrarily narrow, leading to a clustering force that is
very strong in a narrow region, but practically non-existent elsewhere. In
other words, without weight decay, the classifier would overfit to the pair
sums and the embeddings would become stuck in a local minimum.

In the case of alignment, weight decay makes the alignment force (Equation
\ref{eq:alignment}) possible by limiting the magnitude of the biases of the
linear layer, as I show in Appendix \ref{appendix:magnitudes}. This forces the
ReLU activations of the linear layer to remain very close to the origin,
creating the sort of alignment we have seen. Without weight decay, biases could
get arbitrarily large and there would not be any specific point around which
the alignment could take place.

This represents a very interesting link between regularization and training
dynamics. By regularizing one part of the model, we enable the emergence of
desirable training dynamics in another part of the model. Without weight decay,
a layer could overfit to the current state of the other layers, limiting them
from making further progress. This could explain why weight decay is so
effective in training deep learning models
\citep{andriushchenko2023weightdecay}, which consist of many different layers
that need to co-evolve harmoniously during training.

To the best of my knowledge, this link between regularization and training
dynamics has not been previously reported. This opens up interesting new
questions for future research, such as providing theoretical proof for this
phenomenon or showing how it takes place in other problems and architectures.

\newpage
\section{Embedding vectors}
\label{appendix:embeddings}

Below I visualize the embedding vectors that emerge from several random
initializations in the trained transformer for various values of weight decay,
after training for 2000 epochs with a learning rate of 0.01.

\ifthenelse{\boolean{showfigures}}{
    \foreach \w in {1.0, 0.6, 0.3, 0.0} {
            \subsection{$\mathrm{Weight\ decay} = \w$}
            \begin{center}
                \foreach \n in {10,...,29} {
                        \includesvg[height=2.5cm]{svg/embeds_wd\w/\n.svg}
                    }
            \end{center}
            \newpage
        }
}{}

\section{Particle simulation}
\label{appendix:particles}

Below I visualize several particle arrangements that emerge from different
random initializations in the particle simulation for various values of $f_a$,
after running the simulation for 100 steps. I maintain $N = 17, D = 2, g_r =
    g_a = 1$.

\ifthenelse{\boolean{showfigures}}{
    \foreach \f in {1.0, 2.0, 0.5} {
            \subsection{$f_a = \f$}
            \begin{center}
                \foreach \n in {10,...,29} {
                        \includesvg[height=2.5cm]{svg/particles_\f/\n.svg}
                    }
            \end{center}
            \newpage
        }
}{}

\newpage
\section{Classifier visualization}
\label{appendix:classifier}

Below I visualize the embedding vectors (left) and the classification function
formed by the subsequent layers (right) of several models at the end of
training. In the right plot, I also plot the sums of the embedding vectors of
all pairs in the training set (``pair sums''). Models were trained for 2000
epochs with a learning rate of 0.01 and a weight decay of 0.6.

\ifthenelse{\boolean{showfigures}}{
    \begin{center}
        \foreach \n in {10,...,23} {
                \includesvg[height=2.5cm]{svg/color_map/\n.svg}
                \includegraphics[height=2.5cm]{color_map/\n.png}
                \hfill
            }
    \end{center}
}{}

\newpage
\section{Embedding vectors in 3D and 4D}
\label{appendix:multidim}

Below I visualize the embedding vectors that emerge from several random
initializations in the trained transformer with 3-dimensional and 4-dimensional
embeddings. I train for 2000 epochs with a weight decay of 1 and a learning
rate of 0.01.

I use three 2D plots to visualize each configuration of embeddings. I project
the embeddings onto pairs of dimensions and plot the resulting 2D projections.
It is not possible to visualize 3D and 4D embeddings directly, but we can still
get a sense of their grid-like or circular structure by examining the 2D
projections.

\subsection{3-dimensional embeddings}

\ifthenelse{\boolean{showfigures}}{
    \begin{center}
        \foreach \n in {0,...,6} {
                \includesvg[height=2.5cm]{svg/embeds_3d/0\n_0.svg}
                \includesvg[height=2.5cm]{svg/embeds_3d/0\n_1.svg}
                \includesvg[height=2.5cm]{svg/embeds_3d/0\n_2.svg}
                \\
            }
    \end{center}
}{}

\subsection{4-dimensional embeddings}

\ifthenelse{\boolean{showfigures}}{
    \begin{center}
        \foreach \n in {0,...,7} {
                \includesvg[height=2.5cm]{svg/embeds_4d/0\n_0.svg}
                \includesvg[height=2.5cm]{svg/embeds_4d/0\n_1.svg}
                \includesvg[height=2.5cm]{svg/embeds_4d/0\n_2.svg}
                \\
            }
    \end{center}
}{}


\end{document}